\pdfoutput=1
\documentclass[11pt]{article}

\usepackage[]{naacl2021}

\usepackage{times}
\usepackage{latexsym}
\usepackage{graphicx}
\usepackage{amssymb}% http://ctan.org/pkg/amssymb
\usepackage{pifont}% http://ctan.org/pkg/pifont
\newcommand{\cmark}{\ding{51}}%
\newcommand{\xmark}{\ding{55}}%
\usepackage{amsmath}
\usepackage{tabu,multirow}
\usepackage{todonotes}

\usepackage[]{todonotes}

\usepackage[T1]{fontenc}
\usepackage[utf8]{inputenc}

\definecolor{sm}{HTML}{FF0000}
\definecolor{tc}{HTML}{0000FF}

\usepackage{microtype}

\title{MERMAID: Metaphor Generation with Symbolism \\and Discriminative Decoding %MEtaphoR generation with \\syMbolism And dIscriminative Decoding
}

\author{Tuhin Chakrabarty\textsuperscript{1} 
  Xurui Zhang\textsuperscript{3}, 
  Smaranda Muresan\textsuperscript{1,4}
  \textbf{and} \textbf{Nanyun Peng}\textsuperscript{2}\\ 
  \textsuperscript{1}Department of Computer Science, Columbia University\\
  \textsuperscript{2} Computer Science Department, University of California, Los Angeles, \\
  \textsuperscript{3}Tsinghua University,
  \textsuperscript{4}Data Science Institute, Columbia University\\\AND
  {\tt \{tuhin.chakr, smara\}@cs.columbia.edu}\\
  {\tt thuzxr@gmail.com},
  {\tt violetpeng@cs.ucla.edu}
  }

\begin{document}
\maketitle
\begin{abstract}
%\smnote{This sentence needs replacing, particularly since you are working specifically for verbs. I suggest to remove.}
%\smnote{Why not have as title the name of your model? Yuo have the quote in intro, do not put in title. and neural approach is vague.}
%Metaphors are a form of figurative language, in which a word or phrase literally denoting one kind of object or idea is used in place of another to suggest a likeness or analogy between them (as in drowning in money). 
% \violet{you may want to refine the definition here either follow the dictionary definition, e.g., \url{https://www.merriam-webster.com/dictionary/metaphor} or wikipedia definition.}
% Like Symbolism, metaphors also use words to represent an abstract idea. It replaces a literal word with another word that isn't similar and proves they actually have something in common. \violet{I may just remove this sentence, if we have a clear definition of metaphor.}
Generating metaphors is a challenging task as it requires a proper understanding of abstract concepts, making connections between unrelated concepts, and deviating from the literal meaning. In this paper, we aim to generate a metaphoric sentence given a literal expression by replacing relevant verbs. Based on a theoretically-grounded connection between metaphors and symbols, we propose a method to automatically construct a parallel corpus by transforming a large number of metaphorical sentences from the Gutenberg Poetry corpus \cite{jacobs2018gutenberg} to their literal counterpart using recent advances in masked language modeling coupled with commonsense inference. For the generation task, we incorporate a metaphor discriminator to guide the decoding of a sequence to sequence model fine-tuned on our parallel data to generate high quality metaphors. Human evaluation on an independent test set of literal statements shows that our best model generates metaphors better than three well-crafted baselines 66\% of the time on average. Moreover, a task-based evaluation shows that human-written poems enhanced with metaphors proposed by our model are preferred 68\% of the time compared to poems without metaphors. 
%SM-cr we need to say we focus on verbs only no? Here what we say in the ACL paper "In this paper, we aim to generate a metaphoric sentence given a literal expression by replacing relevant verbs. "
%SM should we use competitive baselines rather than well-crafted? 
%humans 21\% of the time and 
%\smnote{We need to say in abstract we work on verbs only.}
\end{abstract}

% Metaphors make our subject more relatable to the reader or to make a complex thought easier to understand. They can also be of tremendous help in enhancing one's writing with imagery

\section{Introduction}
%\smnote{We need better way to introduce metafors, i can do that.}
%SM I think people know what metaphors are I would start bolding with Kundera's quote :)
%A metaphor is a figure of speech that, for rhetorical effect, directly refers to one thing by mentioning another. It employs words in a way that deviates from their normal meaning to represent
%another concept \cite{sporleder2009unsupervised}.
Czech novelist Milan Kundera in his book ``The unbearable lightness of being" said
\begin{center}
    \textit{``Metaphors are not to be trifled with. A single metaphor can give birth to love."}
\end{center}
Metaphors allow us to communicate not just information, but also feelings and complex attitudes \cite{veale2016metaphor}. %SM-cr I removed "real" in front of feelings since it could be fake feelings and can still use metaphors :)
While most computational work has focused on metaphor detection \cite{gao2018neural,stowe2019linguistic,shutova2010metaphor,tsvetkov2014metaphor,veale2016metaphor,stowe2018leveraging}, research on metaphor generation is under-explored \cite{yu-wan-2019-avoid,metagen2}. Generating metaphors could impact many downstream applications such as creative writing assistance, literary or poetic content creation. 

\begin{table}[t]
\centering
\small
\begin{tabular}{|l|l|}
\hline
Literal Input1 & \begin{tabular}[c]{@{}l@{}}The wildfire \textbf{\color{orange}spread} through the forest \\at an amazing speed.\end{tabular}   \\ \hline
GenMetaphor1   & \begin{tabular}[c]{@{}l@{}}The wildfire \textbf{\color{teal}danced} through the forest \\at an amazing speed.\end{tabular}   \\ \hline
Literal Input2 & \begin{tabular}[c]{@{}l@{}}The window panes were \textbf{\color{orange}rattling} as the \\wind blew through them\end{tabular}  \\ \hline
GenMetaphor2   & \begin{tabular}[c]{@{}l@{}}The window panes were \textbf{\color{teal}trembling} as\\ the wind blew through them\end{tabular} \\ \hline
\end{tabular}
\caption{Examples of two generated metaphors GenMetaphor1 and GenMetaphor2 by our best model \textsc{mermaid} from their literal inputs.}
\label{table:example1}
% \vspace{-1em}
\end{table}

Relevant statistics demonstrate that the most frequent type of metaphor is expressed by verbs \cite{steen2010method,martin2006corpus}. We therefore focus on the task of generating a metaphor starting from a literal utterance \cite{metagen2}, where we transform a literal verb to a metaphorical verb. Table~\ref{table:example1} shows examples of literal sentences and the generated metaphors.

%SM I agree, that is what I thought as well
%\violet{I would even propose that we move the challenges here, first talk about challenge, then using symbols will be one solution that we propose to tackle the challenge of lacking parallel data.}

To tackle the metaphor generation problem we need to address three challenges: 1) the lack of training data that consists of pairs of literal utterances and their equivalent metaphorical version in order to train a supervised model; 
2) ensuring that amongst the seemingly endless variety of metaphoric expressions the generated metaphor can fairly consistently capture the same general meaning as the literal one, with a wide variety of lexical variation; and 3) computationally overcome the innate tendency of generative language models to produce literal text over metaphorical one.

%we take advantage of two distinct yet closely related concepts \textit{metaphor} and \textit{symbols}. 

%\smnote{I would first say to solve this problem we need to address this challenges. And then say we propose MERMAID and then the two contributions, they are both part of MERMAID not only the second contriub. I will redo this. }
 
%SM I moved this earlier. 
%There are three main challenges we need to address: 1) the lack of training data that consists of pairs of literal utterances and their equivalent metaphorical version in order to train a supervised model; 
%2) ensuring that amongst the seemingly endless variety of metaphoric expressions the generated metaphor can fairly consistently capture the same general meaning as the literal, with a wide variety of lexical variation) 3) Computationally overcome the innate tendency of generative language models to produce literal text over metaphorical. 

In an attempt to address all these challenges, we introduce our approach for metaphor generation called \textsc{mermaid} ({ME}taphor gene{R}ation with sy{M}bolism {A}nd d{I}scriminative {D}ecoding), making the following contributions:

%\violet{I would avoid using too much boldface.}
%\textbf{Automatically created parallel corpus of \textit{[literal sentence, metaphorical sentence]} pairs}. 
\begin{itemize}
\item{
A method to automatically construct a corpus that contains 93,498 parallel \textit{[literal sentence, metaphorical sentence]} pairs by leveraging the theoretically-grounded relation between \textit{metaphor} and \textit{symbols}. \citet{barsalou1999perceptual} showed how \textit{perceptual symbols} arising from perception are used in conceptual tasks such as representing propositions and abstract concepts. Philosopher Susanne Langer in her essay ``Expressiveness and Symbolism'' stated ``A metaphor is not language, it is an idea expressed by language, an idea that in its turn functions as a \textit{symbol} to express something''. %Looking at the two examples in Table \ref{table:example1}: In the first example, both the GenMetaphor1 evokes the symbolic meaning \textit{power and agility} of ``dancing" to an inanimate object ``wildfire". While in the second example, GenMetaphor2 evokes the symbolic meaning \textit{fear and danger} of ``trembling" to an inanimate object ``window pane". 
Our approach has two steps: 1) identify a set of %\emph{metaphorical} sentences 
sentences that contains \emph{metaphorical verbs} 
from an online poetry corpus; 2) convert these metaphorical sentences to their literal versions using Masked Language Models and structured common sense knowledge achieved from COMET \cite{comet}, a language model fine-tuned on ConceptNet \cite{conceptnet}. For the later, we exploit the SymbolOf relation to make sure the generated sentence that contains the literal sense of the verb has the same symbol as the metaphorical sentence.  %TC Whole sentence is feeded to COMET not the verb only
%SM yes conveyed that, but i had to refer to the fact we still mask verbs so sentence that contain the verbs
%\violet{here I'll add discussions about the relation between metaphor and symbols and draw connections between our steps and the usage of symbolism.}
%\smnote{Replace this example with the ones in Table 1, we already have there 2 examples no need for 3rd.}
%\tcnote{But smara table 1 examples are literal to metaphorical generated by BART , not the distantly labeled generation}
%SM AH, right, nevermind, I might articulate to make it clear. 
For example, for the metaphorical sentence ``The turbulent feelings that \textit{surged} through his soul" our method will generate ``The turbulent feelings that \textit{continued} through his soul" maintaining the common symbolic meaning of \textit{(love, loss, despair, sorrow, loneliness)} between the two (Section \ref{section:data}). }

%\textbf{Transfer learning from a pre-trained model with Discriminative Decoding for generating high quality metaphors.}  
\item{Use of a metaphor discriminator to guide the decoding of a %large-scale 
sequence-to-sequence %language 
model fine-tuned on our parallel data to generate high quality metaphors.
Our system \textsc{mermaid}, \emph{fine-tunes} BART \cite{lewis2019bart} -- a state of the art pre-trained denoising autoencoder built with a sequence to sequence model, on our \emph{automatically collected parallel corpus} of \textit{[literal sentence, metaphorical sentence]} pairs (Sec. \ref{ssection:ft_bart}) to generate metaphors. A discriminative model trained in identifying metaphors is further used to complement our generator and guide the decoding process to improve the generated output (Sec. \ref{ssection:discr}). Human evaluations show that this approach generates metaphors that are better than two literary experts 21\% of the time on average, better 81\% of the time than two well-crafted baselines, and better 36\% of the time than fine-tuned BART \cite{lewis2019bart} (Section \ref{sec:results}). }

\item{A task-based evaluation: improving the quality of human written poems. Evaluation via Amazon Mechanical Turk shows that poems enhanced with metaphors generated by \textsc{mermaid} are preferred by Turkers 68\% of the times compared to poems without metaphors, which are preferred 32\% of the times (Section \ref{sec:task}). 
}
\\
Our code, data and models can be found at \url{https://github.com/tuhinjubcse/MetaphorGenNAACL2021}
\end{itemize}

\section{Dataset Creation with MLM and Symbolism } \label{section:data}

% \violet{since you have more technical details both in the data collection step and in the modeling side, I suggest you to separate the sections for dataset collection and modeling. E.g., you can do ``Dataset Creation with MLM and Symbolism'' and then ``Metaphor Generation Model''}

% Our approach for metaphor generation from literal descriptive sentences has three steps: 1) first convert potential metaphorical sentences from a large scale poetry corpus into literal sentences using Masked Language Models and structured common sense symbolism (Section \ref{section:data}); 2) given the \textit{[literal sentence, metaphorical sentence]} pairs, fine-tune a seq2seq model on these pairs to generate a metaphorical sentence given a literal sentence (Section \ref{section:model}). This two-step approach is shown in the upper half of Figure \ref{figure:sim}.

\begin{figure*}[ht]
\centering
\includegraphics[scale=0.6]{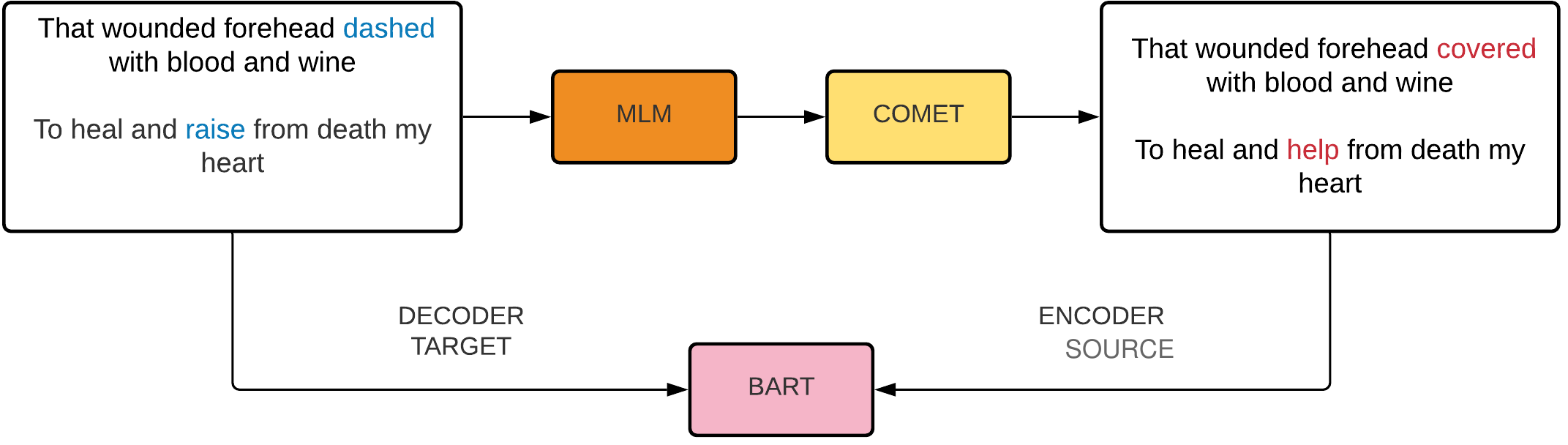}
\caption{\label{figure:sim} A schematic illustration of our system, which shows the \textbf{data creation} and \textbf{training} process where we use MLM along with COMET to transform an original metaphorical input to a literal output evoking similar symbolic meaning and use them to fine-tune BART.}
\vspace{-.5em}
\end{figure*}

% The block below shows the \textbf{decoding} step where we use fine-tuned BART along with a metaphor detecting discriminator to generate a metaphorical sentence conditioned on a literal input

% \subsection{Automatic Parallel Data Creation}
Datasets for metaphors are scarce. To our knowledge, there is no large scale parallel corpora containing literal and metaphoric paraphrases. The closest and most useful work is that of \citet{mohammad-etal-2016-metaphor}. 
However the size of this data-set is small: 171 instances, which is not sufficient to train deep learning models. Recently, \citet{metagen2} rely on available metaphor detection datasets to generate metaphors by a metaphor-masking framework, where they replace metaphoric words in the input texts with metaphor masks (a unique ``metaphor'' token), hiding the lexical item. This creates artificial parallel training data: the input is the masked text, with the hidden metaphorical word, and the output is the original text (e.g., The war [MASK] many people $\rightarrow$ The war \textit{uprooted} many people). The major issue with such masking strategy is that it ignores the semantic mapping between the literal verb and the metaphorical verb. %bottleneck being 
Moreover, there are only 11,593 such parallel instances, still too small to train a neural model.
% metaphoric masks . \smnote{Why is bottleneck?? why do you mean metaphoric masks}
The lack of semantic mapping between the artificial parallel training data samples, coupled with limited size thus affects the lexical diversity and meaning preservation of generated metaphors at test time. In light of these challenges, we propose to compose a  large-scale parallel corpora with literal and metaphorical sentence pairs to learn the semantic mappings. We start with collecting a large-scale corpora of metaphorical sentences (Section \ref{section:metaphor_data}) and leverage masked language model and symbolism-relevant common sense knowledge to create literal version for each metaphorical sentence (Section \ref{section:literal_data}).
%posit the urgency of a real and large scale parallel corpus to train a supervised generative model for metaphor generation. To this end we follow the steps in Section \ref{section:data2} and \ref{section:data2} respectively to automatically create a parallel corpus 
% \subsubsection{Fine-tuning BERT to identify metaphors} \label{section:data1}

% %SM not sure why we need to say that
% %As metaphor identification has been approached as a sequence tagging problem, 
%  \smnote{this section does not makes sense on its own. Merge this with next section on Metaphor data collection. Introduce first the dataset and then say to automatically detect we did.... Refer to Fig 2 or delete Fig2 (it is straightforward} 

% \smnote{and here you briefly say how you built this model or use a paragraph.}

\subsection{Metaphor Dataset Collection} \label{section:metaphor_data}
Metaphors are frequently used in \textit{Poetry} to explain and elucidate emotions, feelings, relationships and other elements that could not be described in ordinary language. We use this intuition to identify a naturally occurring poetry corpus that contains metaphors called Gutenberg Poetry Corpus \cite{jacobs2018gutenberg}.\footnote{\url{https://github.com/aparrish/gutenberg-poetry-corpus}} The corpus contains 3,085,117 lines of poetry extracted from hundreds of books. Not every sentence in the corpus contains a metaphorical verb. So as a first step, we identify and filter sentences containing a metaphorical verb.

\begin{figure}[t]
    \centering
    \includegraphics[scale=0.6]{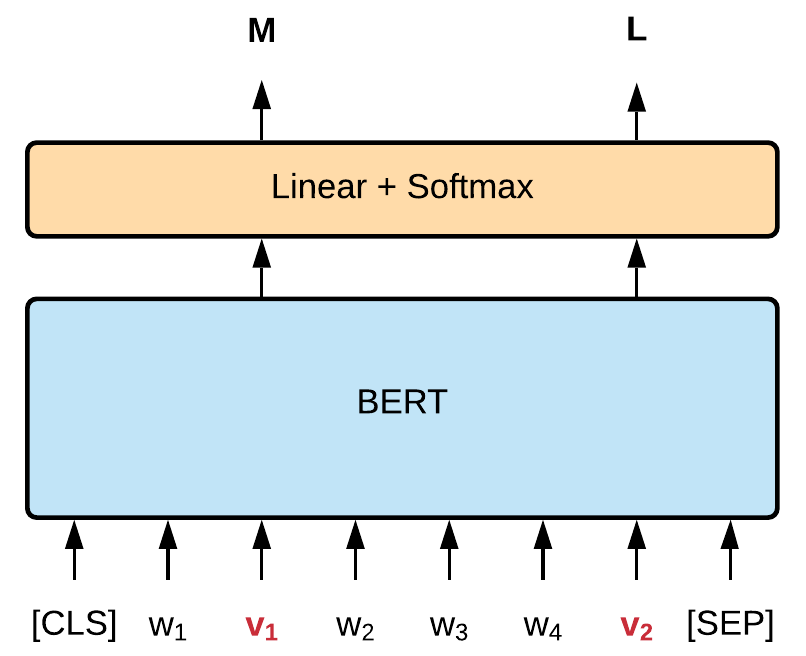}
    \caption{BERT-base-cased model to identify metaphoric verbs, where $v_1$ and $v_2$ represent the verbs in a sentence. (M) denotes softmax probabality of a verb being metaphorical, while (L) denotes it literal softmax probability.}
    \label{figure:detection}
\end{figure}

We build a classifier by fine-tuning BERT~\cite{devlin2018bert} on a metaphor detection corpus \textsc{VU Amsterdam}~\cite{steen2010method}. Since our work is focused on verbs, we only do token classification and calculate loss for verbs. Figure \ref{figure:detection} illustrates the BERT-based token-level classifier. The classification accuracy on test set is 74.7\%, which is on par with most state of art methods.

% \violet{should refer to Figure~\ref{figure:detection}. Also, should add caption to Figure~\ref{figure:detection} to explain that's M and L. Also may worth mention the accuracy of this classifier.}
%To classify literal and metaphorical words we use BERT \cite{devlin2018bert}. Since our work is focused on verbs, we only do token classification and calculate loss for verbs. We use the VU Amsterdam Metaphor Corpus \cite{steen2010method} to fine-tune our detection model. This corpus has 12,542 target verbs as a train set.

Using the metaphor detection model, we identify 622,248 ($20.2\%$) sentences predicted by our model as containing a metaphoric verb. Considering the classifier can introduce noise as the accuracy of the metaphor detection model is far from oracle $100\%$, we only retain sentences which are predicted by our model with a confidence score of $95\%$ (i.e., prediction probability 0.95). This results in a total number of 518,865 ($16.8\%$) metaphorical sentences.

\begin{table}[t]
\centering
\small
\begin{tabular}{|p{0.97cm}|l|}
\hline
Input                                                                        & \begin{tabular}[c]{@{}l@{}}The turbulent feelings that \textbf{surged} \\ through his soul .\end{tabular}                                                                                                                                              \\ \hline\hline
Masked                                                                              & \begin{tabular}[c]{@{}l@{}}The turbulent feelings that {[}MASK{]} \\ through his soul .\end{tabular}                                                                                                                                             \\ \hline\hline
\begin{tabular}[c]{@{}l@{}}Ranked\\ by MLM\\ Prob\end{tabular}        & \begin{tabular}[c]{@{}l@{}}(`tore', 0.11), (`ran', 0.10), (`ripped', 0.09)\\, (`flowed', 0.03), (`rushed', 0.01), .....  , \\(`eased', 0.01),.... , (`continued', 0.0005),...\end{tabular} \\ \hline\hline
\begin{tabular}[c]{@{}l@{}}Ranked\\ by Meta \\ Prob\end{tabular} & \begin{tabular}[c]{@{}l@{}}(`eased', 0.12), (`continued',0.0008), (`spread',\\0.0004), (`kicked', \textbf{0.99}\color{black}{) ,(`punched',}\\ \textbf{0.99}\color{black}){,.....,}(`screamed', \textbf{0.99}\color{black}{),.....}\end{tabular}                    \\ \hline
\end{tabular}
\caption{Table showing a metaphorical sentence (Row1) where the metaphorical verb \textit{surge} is masked (Row2). Row3 shows predicted tokens ranked by default LM probability. Row4 shows predicted tokens ranked by metaphoricity scores obtain from model described in \ref{section:metaphor_data}. Lower scores means more literal.}
\label{table:example2}
\end{table}

%The challenge with this is that the order of predicted tokens from a MLM is based on the default language model probability. 

\subsection{Metaphoric to Literal Transformation with Symbolism} \label{section:literal_data}
After identifying high quality metaphorical sentences, we want to obtain their literal counterparts to create a parallel training data. Masked language models like BERT \cite{devlin2018bert}, or roBERTa \cite{liu2019roberta} can be used for fill-in-the-blank tasks, where the model uses the context words surrounding a masked token to predict the masked word. We borrow this framework to mask the metaphorical verb (Table \ref{table:example2} Row1 vs Row2) from a sentence and use BERT-base-cased model to obtain the top $200$ candidate verbs to replace the metaphorical one to generate literal sentences (Table \ref{table:example2} Row3). There are two main issues in solely relying on MLM predicted verbs: 1) they are not necessarily literal in nature; 2) after replacing the default MLM predicted verb, the metaphorical sentence and the new sentence with the replaced verb might be semantically dissimilar.

\subsubsection{Ensuring Literal Sense}Even though our inductive biases tell us that the chance of a predicted token having a literal sense is higher than having a metaphorical one, this cannot be assumed. To filter only literal candidate verbs we re-rank the MLM predicted mask tokens based on literal scores obtained from \ref{section:metaphor_data} since the model can predict the softmax probability of a verb in a sentence being either literal or metaphorical (Table \ref{table:example2} Row4).

% \smnote{I will work to rephrase this below it is too wordy. }
\subsubsection{Ensuring Meaning Preservation}
While we can potentially pair the sentence with the top most literal ranked verb with the input sentence containing the metaphorical verb, they might symbolically or semantically represent different abstract concepts. For example, in Table \ref{table:example3}, after replacing the metaphorical verb ``surge" with the top most literal verb ``eased", the sentence \textit{``The turbulent feelings that eased through his soul"} evoke a different symbolic meaning of \textit{peace,love,happiness,joy \& hope} in comparison to the input containing the metaphorical verb, which evokes a symbolic meaning of \textit{love, loss, despair, sorrow \& loneliness}. To tackle this problem we ensure that the transformed literal output represents the same symbolic meaning as the metaphorical input.

% While the symbolic meanings are naturally known to humans through common sense and connotative knowledge, computers still struggle to perform well on such tasks.
To generate the common sense {\tt SYMBOL} that is implied by the literal or metaphorical sentences, we feed the sentences as input to COMET \cite{comet} and restrict it to return top-5 beams. %COMET is an adaptation framework for constructing commonsense knowledge based on pre-trained language models. 
%SM-cr I replace the above sentence since it is weired. 
COMET is an adapted knowledge model pre-trained on ConceptNet.\footnote{\url{https://mosaickg.apps.allenai.org/comet\_conceptnet}}  Our work only leverages the \textbf{SymbolOf} relation from COMET.

\begin{table}[t]
\centering
\small
\begin{tabular}{|l|l|l|}
\hline
Meta Input                                                 & \multicolumn{2}{l|}{\begin{tabular}[c]{@{}l@{}}The turbulent feelings that \textbf{surged} \\through his soul .\end{tabular}} \\ \hline
\begin{tabular}[c]{@{}l@{}}Inp Symbol\end{tabular} & \multicolumn{2}{l|}{love, loss, despair, sorrow, loneliness}                                                         \\ \hline
Lit Output1                                               & \begin{tabular}[c]{@{}l@{}}The turbulent feelings that\\ \textit{\color{red}eased} through his soul .\end{tabular}             &  \color{red}{\xmark \par}       \\ \hline
Symbol                                                 & \multicolumn{2}{l|}{peace,love,happiness,joy,hope}                                                                   \\ \hline
Lit Output2                                              & \begin{tabular}[c]{@{}l@{}}The turbulent feelings that\\ \textit{\color{blue}continued} through his soul .\end{tabular}         &   \color{blue}{\cmark}       \\ \hline
Symbol                                                 & \multicolumn{2}{l|}{love, loss, despair, sorrow, loneliness}                                                         \\ \hline
% Meta Inp2                                                  & \multicolumn{2}{l|}{\begin{tabular}[c]{@{}l@{}}Who \textbf{fishes} for the truth and wanteth \\skill .\end{tabular}}          \\ \hline
% \begin{tabular}[c]{@{}l@{}}Inp Symbol\end{tabular} &  \multicolumn{2}{l|}{\begin{tabular}[c]{@{}l@{}}knowledge, cleverness,  intelligence,\\ truth, honesty\end{tabular}}  \\ \hline
% Literal Op3                                              & \begin{tabular}[c]{@{}l@{}}Who \textit{\color{red}fell} for the truth and\\ wanteth skill .\end{tabular}                       &  \color{red}{\xmark \par}       \\ \hline
% Symbol                                                 & \multicolumn{2}{l|}{\begin{tabular}[c]{@{}l@{}}cleverness, truth, honesty lie, deception\end{tabular}}             \\ \hline
% Literal Op4                                               & \begin{tabular}[c]{@{}l@{}}Who \textit{\color{blue}intended} for the truth and\\ wanteth skill .\end{tabular}                   &  \color{blue}{\cmark}       \\ \hline
% Symbol                                                 & \multicolumn{2}{l|}{\begin{tabular}[c]{@{}l@{}}knowledge, cleverness, intelligence,\\ truth, honesty\end{tabular}}  \\ \hline
\end{tabular}
\caption{Table showing input metaphorical sentence and literal outputs along with the associated symbolic meaning obtained from COMET \cite{comet}. Lit Output1 is an incorrect candidate since the symbolic meanings are divergent.}
\label{table:example3}
\end{table}
% \smnote{What two resources?}
%SM do you mean below symbolic is symbolism-related? 
We now need a method to combine information from  MLM and symbolic knowledge obtained from COMET described above. To do this, we filter candidates from MLM token predictions based on the symbolic meaning overlap between the metaphorical input and literal output first. To ensure that the quality is high, we put a strict requirement that all the 5 symbolic beams (typically words or short phrases) for the input metaphorical sentence should match all the 5 symbolic beams for the output literal sentence. Between multiple literal candidates all having beam overlap of 5, they are further ranked by reverse metaphoricity (i.e., literal) scores. The top most candidate is returned thereafter. 
We finally end up with 90,000 pairs for training and 3,498 pairs for validation.

\section{Metaphor Generation} \label{section:model}

\begin{figure*}[ht]
\centering
\includegraphics[scale=0.6]{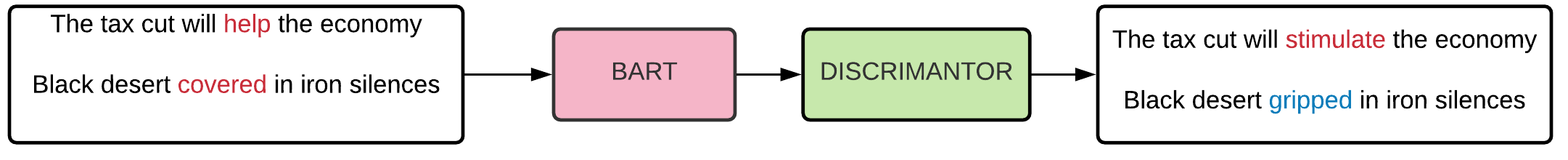}
\caption{\label{figure:sim1} Schematic showing the \textbf{decoding} step where we use fine-tuned BART along with a metaphor detecting discriminator to generate a metaphorical sentence conditioned on a literal input}
\vspace{-.5em}
\end{figure*}

% \smnote{See my suggestion to rename this if the entire approach is MERMAID not only this part}
Our goal of generating metaphors can be broken down into two primary tasks: 1) generating the appropriate substitutions for the literal verb while being pertinent to the context; 2) ensuring that the generated utterances are actually metaphorical.

\subsection{Transfer Learning from BART} \label{ssection:ft_bart}
To achieve the first goal, we fine-tune BART \cite{lewis2019bart}, a pre-trained conditional language model that combines bidirectional and auto-regressive transformers, on the collected parallel corpora. %It is implemented as a sequence-to-sequence model. 
%Here, the encoder input is a sequence of words, and the decoder generates outputs auto-regressively. We refer the reader to \cite{lewis2019bart} for further details. 
Specifically, we fine-tune BART by treating the literal input as encoder source and the metaphorical output as the the decoder target (Figure \ref{figure:sim}). One issue of the pre-trained language models is that they have a
tendency to generate literal tokens over metaphorical ones. To overcome this, we introduce a rescoring model during the decoding process to favor more metaphorical verbs. The rescoring model is inspired by \citet{holtzman2018learning,goldfarb2020content} and detailed in the next section.

\subsection{Discriminative Decoding} \label{ssection:discr}
We have a base metaphor generation model $p(\textbf{z}|\textbf{x})$ which is learned by fine-tuning BART~\cite{lewis2019bart} on pairs of literal ($x$) and metaphorical ($z$) sentences. 
%We posit that a model further trained on discriminating between literal versus metaphorical sentences would be ideal for guiding our decoding process. Hence, 
We propose to modify the decoding objective to incorporate a Metaphor detection rescoring model $a$ and re-rank the base, or ``naive" BART generated hypotheses, bringing the metaphoric representation closer to the rescoring model's specialty and desirable attribute. The modified decoding objective becomes:

\vspace{-0.5cm}
{\small
\begin{align} \label{eq:decoding}
f_\lambda({\bf x, z}) = \sum_{i}^{m}-\log p(z|z<i,{\bf x}) + \lambda a({\bf x}, z_{i...m}) 
\end{align}
\vspace{-0.5cm}
}\\
where $\lambda$ is a weight of the score given by $a$. 

\paragraph{Implementation Details} \label{sec:imp}
We use top-k sampling strategy \cite{fan2018hierarchical} (k=5) to generate metaphors conditioned on a literal input.
Our rescoring model $a$ is a RoBERTa model fine-tuned on a combined dataset of \cite{steen2010method,beigman-klebanov-etal-2018-corpus} to classify sentences as literal or metaphorical based on whether there exists a metaphorical verb. It is a sentence level task where the model predicts a sentence as literal or metaphorical. We down-sample the data to maintain a ratio of ($1:1$) between two classes and use $90\%$ of the data to train and $10\%$ for validation. We achieve a considerably decent validation accuracy of $83\%$. We manually tune $\lambda$ using grid search on a small subset of 3,498 validation samples from our parallel automatic data and choose the best value.

% \violet{please add  details about how do you choose the  $\lambda$? Did you train it, or manually  tuned it by grid search, or set it empirically?}

Figure \ref{figure:sim1} shows the process of re-ranking BART hypothesis using the discriminator described above to generate novel metaphorical replacements for literal verbs.
All the hyper-parameters for data creation, fine-tuning and discriminative decoding are exactly the same as mentioned in \textbf{Appendix A}.
%SM-cr remove supplemntary material and give appendix number

%One of the questions which comes across is 
The reason to use a separate discriminator for
decoding instead of using the same BERT based classifier used for parallel data creation, %This 
was %particularly done 
to avoid introducing dataset biases or spurious correlations. The BERT-based classifier used for automatically creating the parallel dataset ideally has already picked up salient metaphorical phenomena in the VUA dataset. To further guide the decoding process, we hypothesize that a model trained on datasets not seen during training would lead to better generalization. We  experimented with using the BERT model trained on VUA for rescoring, but the results were not better. 

% for more details about the best values for hyper-parameters.

% \smnote{This section does not refer back to Architecture Figure. }
% \tcnote{Not sure what to refer, maybe u can change however u like smara} 
% \violet{I'm thinking, maybe we can split the current Figure 1 into two figures, one just for dataset creation and another just for generation (and we only need to show the inference time, because the training is somewhat straightforward). This way we can refer to the 2nd figure in this seciotn.}

% Post fine-tuning at the inference step, we use top-k sampling strategy \cite{fan2018hierarchical} to generate similes conditioned on a test literal input.
\section{Experimental Setup}
To compare the quality of the generated metaphors, we benchmark our \textsc{mermaid} model against human performance (i.e., the two creative writing experts HUMAN1 (a novelist) \& HUMAN2 (a poet) who are not the authors of the paper) (Section \ref{testdata}) and three baseline systems described below.

% \violet{I think it's worth re-explain the two experts here. Readers probably have already forget them at this point.} 

\subsection{Baseline Systems} 

\textbf{Lexical Replacement (LEXREP)}: We use the same idea as our data creation process (Section \ref{section:literal_data}). We use our model described in Section \ref{section:metaphor_data} to re-rank the predicted tokens from a mask language model based on metaphoricity scores. We filter the top 25 ranked metaphorical candidates and further rerank them based on symbolic meaning overlap with the literal meaning using COMET \cite{comet} and replace the literal verb with the top scoring candidate.

\textbf{Metaphor Masking (META\_M)}: We use the metaphor masking model proposed by \citet{metagen2} where the language model learns to replace a masked verb with a metaphor. They train a seq2seq model with the encoder input of the format (The tax cut [MASK] the economy) and the decoder output being the actual metaphorical sentence (The tax cut lifted the economy). During inference, they mask the literal verb and expect the language model to infill a metaphorical verb.

\textbf{BART}: We use generations from a BART model fine-tuned on our automatically created data without the discriminative decoding. This helps us gauge the effect of transfer learning from a large generative pre-trained model, which also accounts for context unlike the retrieval based methods.

\subsection{Test Data} \label{testdata}
To measure the effectiveness of our approach, we need to evaluate our model on a dataset that is independent of our automatically created parallel data and that is diverse across various domains, genres and types. Hence we rely on test data from multiple sources. As our first source, we randomly sample literal and metaphorical sentences with high confidence ($>0.7$) and unique verbs from the existing dataset introduced by \citet{mohammad-etal-2016-metaphor}. For the metaphorical sentences from \citet{mohammad-etal-2016-metaphor} we convert them to their literal equivalent the same way as discussed in Section \ref{section:literal_data} without the use of COMET as we do not need it. To ensure diversity in genre, as our second source we scrape \textsc{WritingPrompt} and \textsc{Ocpoetry} subreddits for sentences with length up to 12 words, which are \emph{literal} in nature based on prediction from our model described in Section \ref{section:metaphor_data}. We collate 500 such sentences combined from all sources and randomly sample 150 literal utterance for evaluation.

We use two literary experts (not authors of this paper) ---  a student in computer science who is also a poet, and a student in comparative literature who is the author of a novel --- to write corresponding metaphors for each of these 150 inputs for evaluation and comparison.

\subsection{Evaluation Criteria}
% \smnote{Why are evaluation criteria part of Approach Section/MERMAID and not Experimental Setup. Move to section 5. }
\paragraph{Automatic evaluation.}
One important aspect in evaluating the quality of the generated metaphors is whether they are faithful to the input: while we change literal sentences to metaphorical ones, it should still maintain the same denotation as the input. To this end, we calculate the \textit{Semantic Similarity} between the metaphorical output and the input using sentence-BERT (SBERT) \cite{reimers2019sentence}. We also calculate corpus-level BLEU-2 \cite{BLEU} and BERTScore \cite{zhang2019bertscore} with human written references.

\paragraph{Human evaluation.}
% \smnote{Hm, you did not evaluate meaning using human eval? I do not trust only automatic metrics for keeping the denotative meaning.}
% \smnote{What 4 systems, you did not introduced the baselines.See my suggestion how to structure experimental setup}
Since automatic evaluation is known to have significant limitations for creative generation~\cite{novikova-etal-2017-need}, we further conduct human evaluation on a total of 900 utterances, 600 generated from 4 systems and 300 generated by the two human experts.  We propose a set of four criteria to evaluate the generated output: (1) \textit{Fluency (Flu)} (``How fluent, grammatical, well formed and easy to understand are the generated utterances?''),
(2) \textit{Meaning (Mea)} (``Are the input and the output referring or meaning the same thing?") (3) \textit{Creativity (Crea)} (``How creative are the generated utterances?''), and (4) \textit{Metaphoricity (Meta)} (``How metaphoric are the generated utterances''). The human evaluation is done on the Amazon Mechanical Turk platform. 
Each Turker was given a literal input and 6 metaphorical outputs (4 from system outputs -- 3 baselines and our proposed system \textsc{mermaid}, and 2 from humans) at a time, with the metaphorical outputs randomly shuffled to avoid potential biases. Turkers were instructed to evaluate the quality of the metaphorical sentences with respect to the input and not in isolation. 
As we evaluate on four dimensions for 900 utterances, we have a total of 3600 evaluations. %We hired Turkers on MTurk to rate . 
Each criteria was rated on a likert scale from 1 (not at all) to 5 (very). Each group of utterances was rated by three separate Turkers, resulted in 42, 48, 44 and 53 Turkers for the four evaluation tasks respectively. We pay them at a rate of \$15 per hour.

\begin{table}[t]
\small
\centering
\begin{tabular}{|l|l|l|l|}
\hline
 \textbf{System}       & \textbf{Similarity $\uparrow$} & \textbf{BLEU-2$\uparrow$} & \textbf{BertScore$\uparrow$} \\ \hline
LEXREP  &      79.6      &  \textbf{68.7} & 0.56            \\ \hline
META\_M &      73.2     &  61.0    & 0.62         \\ \hline
BART    &      83.6      &  65.0  & \underline{0.65}           \\ \hline
\textsc{mermaid} &      \underline{85.0}      &    \underline{66.7}  & \textbf{0.71}         \\ \hline\hline
HUMAN1  &      \textbf{86.6}      &   -    & -        \\ \hline
HUMAN2  &      84.2     &      -      & -   \\ \hline
\end{tabular}
\caption{Automatic evaluation results on test set where \textsc{mermaid} significantly outperforms other automatic methods for 2 out of 3 metrics ($p<.001)$ according to approximate randomization test). BLEU-2 and BertScore is calculated w.r.t to Human references (HUMAN1 \& HUMAN2). Corpus level BLEU-2 and Semantic Similarity are in range of (0-100) while BertScore is in range (0-1) }
\label{table:auto}
\end{table}

% \violet{what are you comparing during the significance test? ours and other automatic methods?}

% \smnote{I would restructure this section to have 5.1 System Baselines, 5.2 Human Generated Data (your test data section goes here) 5.3 Evaluation Criteria. If you agree I can do that}
% \smnote{And you can keep this as intro but say first 3 baselines (5.1), then human performance (5.2) and then using evaluation criteria (5.3)}

\section{Results} \label{sec:results}
\begin{table}[]
\small
\centering
\begin{tabular}{|@{ }p{2.5cm}@{ }|@{}p{1.1cm}@{}|@{}p{1.1cm}@{}|@{ }p{1.1cm}@{}|@{ }p{1.1cm}@{}|}
\hline
\bf System & \bf Flu & \bf Mea & \bf Crea & \bf Meta \\ \hline
HUMAN1 & \textbf{3.83}  & \textbf{3.77} &\textbf{4.02}    &  \textbf{3.52}   \\ \hline
HUMAN2 &  3.29  & 3.43 & 3.58   & 3.16  \\ \hline\hline
LEXREP    &  2.21 & 2.59 &  2.16   & 1.98  \\ \hline
META\_M  &  2.10  & 1.91 &  2.00   & 1.89 \\ \hline
BART &  3.33  & 3.08 & 3.16   & 2.85  \\ \hline
\textsc{mermaid}   & \underline{3.46}   & \underline{3.35}&\underline{3.50}    & \underline{3.07}  \\ \hline
\end{tabular}
\caption{Human evaluation on four criteria of metaphors quality for systems and humans generated metaphors. We show average scores on a likert scale of 1-5 where 1 denotes the worst and 5 be the best. Boldface denotes the best results overall and underscore denotes the best among computational models.}
\label{table:example4}
\vspace{-2em}
\end{table}

\begin{table*}[!ht]
\small
\centering
\begin{tabular}{|@{ }p{2.5cm}@{ }|l@{ }|@{ }p{8.5cm}@{ }|@{ }l@{ }|@{ }l@{ }|@{ }l@{ }|@{ }l@{ }|}
\hline
Literal & System & Metaphor & Flu & Mea & Crea & Meta   \\ \hline
\multirow{6}{*}{\begin{tabular}[c]{@{}p{2.5cm}@{}} The scream \textit{filled} the night \end{tabular}} & HUMAN1  & \begin{tabular}[c]{@{}l@{}}  The scream \textit{pierced} the night   \end{tabular}  & \textbf{4.3} & \textbf{5.0} &\textbf{3.7}  & \textbf{4.0}  \\ 
    \cline{2-7}
    & HUMAN2  & \begin{tabular}[c]{@{}l@{}}  The scream \textit{covered} the night  \end{tabular}  & 2.7& 4.0 &3.0 & 3.0 \\
    \cline{2-7}
    & LEXREP & \begin{tabular}[c]{@{}l@{}}  The scream \textit{held} the night \end{tabular}  & 1.7 & 3.7 & 2.0 & 1.7 \\
    \cline{2-7}
    & META\_M & \begin{tabular}[c]{@{}l@{}}  The scream \textit{opened} the night \end{tabular}  & 1.0 & 1.0 & 1.0 & 1.0  \\
    \cline{2-7}
    & BART & \begin{tabular}[c]{@{}l@{}}  The scream \textit{filled} the night \end{tabular}  & 2.3  & 1.0 & 2.3  & 1.0 \\
    \cline{2-7}
    & \textsc{mermaid} & \begin{tabular}[c]{@{}l@{}}  The scream \textit{pierced} the night \end{tabular}  & \textbf{4.3} & \textbf{5.0} & \textbf{3.7} & \textbf{4.0}  \\
    \cline{2-7}
    \hline\hline
\multirow{6}{*}{\begin{tabular}[c]{@{}p{2.5cm}@{}} The wildfire \textit{spread} through the forest at an amazing speed
 \end{tabular}} & HUMAN1 & \begin{tabular}[c]{@{}l@{}} The wildfire \textit{ravaged} through the forest at an amazing speed  \end{tabular}  & \textbf{4.7}  & \textbf{4.3} & 4.0 &  3.0 \\ 
    \cline{2-7}
    & HUMAN2 & \begin{tabular}[c]{@{}l@{}} The wildfire \textit{leapt} through the forest at an amazing speed \end{tabular}  & 3.7 & 3.0 & \textbf{5.0} & 3.7  \\
    \cline{2-7}
    & LEXREP & \begin{tabular}[c]{@{}l@{}} The wildfire \textit{saw} through the forest at an amazing speed \end{tabular}  & 1.3  & 1.0 & 2.7 & 3.3  \\
    \cline{2-7}
    & META\_M & \begin{tabular}[c]{@{}l@{}} The wildfire \textit{grows} through the forest at an amazing speed \end{tabular}  & 3.7 & 2.7 & 2.7 & \textbf{4.0}  \\
    \cline{2-7}
    & BART & \begin{tabular}[c]{@{}l@{}} The wildfire \textit{swept} through the forest at an amazing speed  \end{tabular}  & 4.0 & 3.7 & 4.7  & \textbf{4.0}  \\
    \cline{2-7}
    & \textsc{mermaid} & \begin{tabular}[c]{@{}l@{}} The wildfire \textit{danced} through the forest at an amazing speed \end{tabular}  & 3.0 & 4.0 & 4.0  & 3.7  \\
    \cline{2-7}
    \hline\hline
\multirow{6}{*}{\begin{tabular}[c]{@{}p{2.2cm}@{}} My heart \textit{beats} when he walks in the room
 \end{tabular}} & HUMAN1 & \begin{tabular}[c]{@{}l@{}}My heart \textit{skips} when he walks in the room   \end{tabular}  & 4.7 & \textbf{5.0} & 4.0  & \textbf{4.3}  \\ 
    \cline{2-7}
    & HUMAN2 & \begin{tabular}[c]{@{}l@{}}  \end{tabular}My heart \textit{sings} when he walks in the room  & \textbf{5.0} & 4.3 & 3.7 & 3.3  \\
    \cline{2-7}
    & LEXREP & \begin{tabular}[c]{@{}l@{}}  \end{tabular}My heart \textit{made} when he walks in the room  & 1.0 & 1.0 & 1.0  & 1.0  \\
    \cline{2-7}
    & META\_M & \begin{tabular}[c]{@{}l@{}}  \end{tabular} My heart \textit{came} when he walks in the room & 1.7 & 1.0 & 1.3 & 1.3  \\
    \cline{2-7}
    & BART & \begin{tabular}[c]{@{}l@{}}  \end{tabular} My heart \textit{sings} when he walks in the room &\textbf{5.0} & 4.3 & 3.7 & 3.7  \\
    \cline{2-7}
    & \textsc{mermaid} & \begin{tabular}[c]{@{}l@{}}  \end{tabular} My heart \textit{jumps} when he walks in the room & 4.7  & 4.7 & \textbf{4.3} & 4.0  \\
    \cline{2-7}
    \hline\hline
\multirow{6}{*}{\begin{tabular}[c]{@{}p{2.5cm}@{}} After a glass of wine, he \textit{relaxed} up a bit
 \end{tabular}} & HUMAN1  & \begin{tabular}[c]{@{}l@{}} After a glass of wine, he \textit{loosened} up a bit  \end{tabular}  & \textbf{4.7}  & \textbf{5.0}& \textbf{5.0} & \textbf{4.0}   \\ 
    \cline{2-7}
    & HUMAN2 & \begin{tabular}[c]{@{}l@{}} After a glass of wine, he \textit{unfurled} up a bit \end{tabular}  & 2.0 & \textbf{5.0} & 2.0 & 3.7  \\
    \cline{2-7}
    & LEXREP & \begin{tabular}[c]{@{}l@{}} After a glass of wine, he \textit{followed} up a bit \end{tabular}  &3.7  & 1.0 & 2.7 & 1.7  \\
    \cline{2-7}
    & META\_M & \begin{tabular}[c]{@{}l@{}} After a glass of he \textit{touched} up a bit
 \end{tabular}  &1.3  &1.0 & 1.7  & 2.0  \\
    \cline{2-7}
    & BART & \begin{tabular}[c]{@{}l@{}} After a glass of wine, he \textit{dried} up a bit \end{tabular}  & 2.7  &1.0 & 2.3 &  2.0 \\
    \cline{2-7}
    & \textsc{mermaid} & \begin{tabular}[c]{@{}l@{}} After a glass of wine, he \textit{loosened} up a bit \end{tabular}  & 4.3 & \textbf{5.0}& \textbf{5.0} &  3.7 \\
    \cline{2-7}
    \hline\hline
\multirow{6}{*}{\begin{tabular}[c]{@{}p{2.5cm}@{}} The tax cut will \textit{help} the economy
 \end{tabular}} & HUMAN1 & \begin{tabular}[c]{@{}l@{}} The tax cut will \textit{uplift} the economy  \end{tabular}  & 4.7 & \textbf{5.0} & \textbf{4.7} & \textbf{4.0}  \\ 
    \cline{2-7}
    & HUMAN2 & \begin{tabular}[c]{@{}l@{}}  The tax cut will \textit{fertilize} the economy\end{tabular}  & 4.0  & 4.3 & 4.3 & 3.7  \\
    \cline{2-7}
    & LEXREP & \begin{tabular}[c]{@{}l@{}}  The tax cut will \textit{bring} the economy \end{tabular}  & 1.7  & 3.0 & 2.7 & 1.7  \\
    \cline{2-7}
    & META\_M & \begin{tabular}[c]{@{}l@{}}  The tax cut will \textit{prevent} the economy \end{tabular}  & 1.7 & 1.0 & 2.0 & 1.0  \\
    \cline{2-7}
    & BART & \begin{tabular}[c]{@{}l@{}}  The tax cut will \textit{strengthen} the economy  \end{tabular}  & \textbf{5.0} & \textbf{5.0} & 4.3 & 3.7  \\
    \cline{2-7}
    & \textsc{mermaid} & \begin{tabular}[c]{@{}l@{}}  The tax cut will \textit{stimulate} the economy  \end{tabular}  & \textbf{5.0} & 4.7 & 3.7 & \textbf{4.0}  \\
    \cline{2-7}
    \hline\hline
    \multirow{6}{*}{\begin{tabular}[c]{@{}p{2cm}@{}} I tried to \textit{resolve} things over between them
 \end{tabular}} & HUMAN1 & \begin{tabular}[c]{@{}l@{}} I tried to \textit{tide} things over between them  \end{tabular}  & 4.3 & 3.0 & 3.7  & \textbf{4.3}  \\ 
    \cline{2-7}
    & HUMAN2 & \begin{tabular}[c]{@{}l@{}} I tried to \textit{patch} things over between them \end{tabular}  & \textbf{4.7}& \textbf{4.7} &\textbf{5.0} &  2.0 \\
    \cline{2-7}
    & LEXREP & \begin{tabular}[c]{@{}l@{}} I tried to \textit{push} things over between them \end{tabular}  & 3.3 & 1.0 & 2.3  & 2.0  \\
    \cline{2-7}
    &  META\_M & \begin{tabular}[c]{@{}l@{}} I tried to \textit{make} things over between them  \end{tabular}  & 4.0 & 1.0 & 2.7  & 2.7  \\
    \cline{2-7}
    & BART  & \begin{tabular}[c]{@{}l@{}} I tried to \textit{put} things over between them  \end{tabular}  & \textbf{4.7} & 2.0 & 3.0  & 2.7  \\
    \cline{2-7}
    &  \textsc{mermaid} & \begin{tabular}[c]{@{}l@{}} I tried to \textit{smooth} things over between them  \end{tabular}  & \textbf{4.7} & \textbf{4.7} & \textbf{5.0} & 4.0  \\
    \cline{2-7}
    \hline
\end{tabular}
\caption{Examples of generated outputs from different systems (with human written metaphors as references). We show average scores (over three annotators) on a 1-5 scale with 1 denotes the worst and 5 be the best. The italics texts in the literal column represent the \textit{verb} while those in Metaphor column represents the generated metaphorical \textit{verb}. Boldface indicates the best results.}
\label{table:example5}
\end{table*}

Based on the semantic similarity metric shown in column 1 of Table ~\ref{table:auto}, our system \textsc{mermaid} is better in preserving the meaning of the input than the other baselines. As mentioned, we calculate BLEU-2 and BERTScore between system outputs and human references.
\textsc{mermaid} is better than the other baselines according to BERTScore. In terms of BLEU-2, \textsc{mermaid} is second best.
% \smnote{Did you calculate BLEU and BERTscore betwen HUMANS?}
% \smnote{Semantic similarity based on automatic metric is Ok but i would have like to see a meaning preservation criteria in human eval. i think reviewers will ask}.

Table ~\ref{table:example4} shows the average scores for the human evaluation on four metaphor quality criteria for \textsc{mermaid}, the baselines, and human written metaphors on the test set. The inter-annotator agreements computed using Krippendorff's alpha for Creativity, Meaning, Fluency and Metaphoricity are 0.44, 0.42, 0.68, 0.52 respectively. The results demonstrate that \textsc{mermaid} is significantly better than the baselines on all four criteria ($p<.001$ according to approximate randomization test).

Table \ref{table:example5} presents several generation outputs from different systems along with human judgements on individual criteria. We observe that incorporating a discriminator often guides our model to generate better metaphors than the already strong baseline using BART. Finally, incorporating symbolic meaning in data creation step helps our model to maintain the same meaning as the input.

\section{Task Based Evaluation} \label{sec:task}
\begin{figure}[ht]
\centering
\includegraphics[scale=0.15]{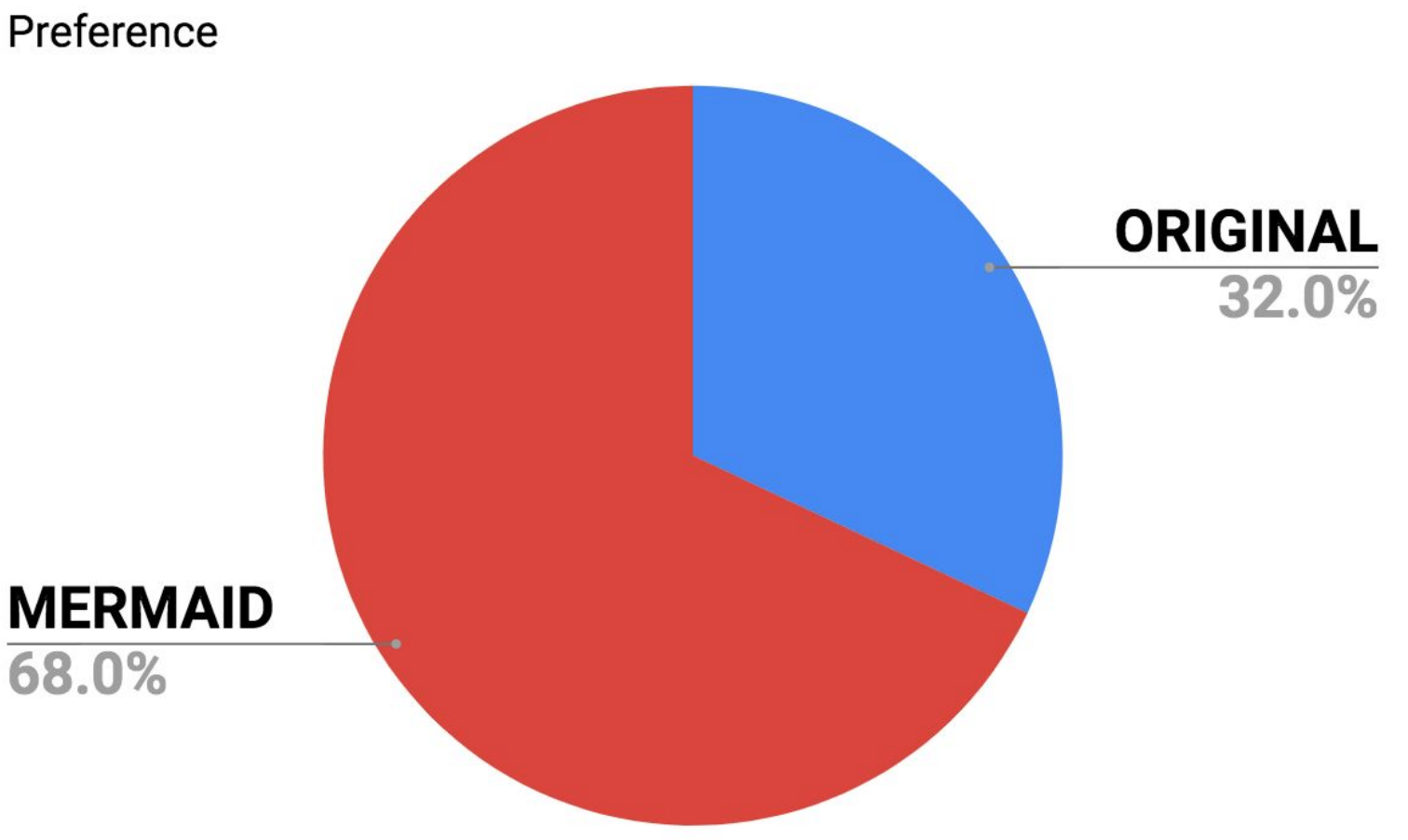}
\caption{\label{figure:poetrypref} Percentage of Preference of Original Quatrains  vs Quatrains rewritten by \textsc{mermaid}}
\vspace{-.5em}
\end{figure}
Metaphors are frequently used by creative writing practitioners, in particular poets, to embellish their work. We posit that \textsc{mermaid} can be used to edit literal sentences in poems to further  enhance creativity. To test this hypothesis, we first crawl original poems submitted by authors from the sub-reddit \textsc{OCPoetry}. The poems are of variable lengths, so to ensure parity we 
break them into \textit{Quatrains} (four sentence stanza). We randomly sample 50 such Quatrains containing at least one sentence with a literal verb in it. We use our metaphor detection model (Section \ref{section:metaphor_data}) to detect literal verbs.

We then  select a sentence containing a literal verb from each Quatrain and use \textsc{mermaid} to re-write it so that the resulting output is metaphorical. We ignore common verbs like \textit{is,was,are,were,have,had}. If there are more than one sentence in Quatrain with literal verbs, we choose the sentence with a literal verb that has the highest probability for being literal. For sentences with multiple literal verbs, we choose the verb with highest literal probability.

\begin{table}[]
\small
\centering
\begin{tabular}{|l|}
\hline
\begin{tabular}[c]{@{}l@{}}And the hills have a shimmer of light between,\\ And the valleys are \textit{\color{red}covered} with misty veils,\\ And .........,\\ ..... \\\end{tabular} \\ \hline
\begin{tabular}[c]{@{}l@{}}And the hills have a shimmer of light between,\\ And the valleys are \textit{\color{blue}wrapped} with misty veils,\\ And .........,\\ \end{tabular} \\ \hline
\begin{tabular}[c]{@{}l@{}}Leaves on a maple, \textit{\color{red}burst} red with the shorter days;\\ Falling to the ground.\\....\\\end{tabular} \\ \hline
\begin{tabular}[c]{@{}l@{}}Leaves on a maple, \textit{\color{blue}burgeoned} red with the shorter days;\\ Falling to the ground. \\....\\\end{tabular} \\ \hline
\end{tabular}
\caption{Example Quatrains from reddit where \textsc{mermaid} rewrites a sentence containing a literal verb to make it metaphorical. }
\label{table:poetryexample}
\end{table}

Our goal is to see if re-written poems are qualitatively better than the original forms. To do this, we hire Turkers from Amazon Mechanical Turk and present them with hits where the task is to choose the better version between the original Quatrain and the re-written version. 15 Turkers were recruited for the task. Each Quatrain was evaluated by 3 distinct Turkers. Table \ref{table:poetryexample} shows metaphorical transformations by a \textsc{mermaid}
Figure \ref{figure:poetrypref} shows that poems rewritten by \textsc{mermaid} were considered better by the Turkers.  

%SM-cr We do not refer to Table 7 of examples. 

 \section{Related Work}
Most researchers focused on identification and interpretation of metaphor, while metaphor generation is relatively under-studied. % new. 
\subsection{Metaphor Detection}
% Several works based on statistical machine learning were proposed in recent decades.
For metaphor detection, researchers focused on variety of features, including unigrams, imageability, sensory features, WordNet, bag-of-words features \cite{klebanov2014different, tsvetkov2014metaphor, shutova2016black, tekirouglu2015exploring, hovy2013identifying, koper2016distinguishing}.
% These works explicitly exploited linguistic cognition.
\par
% With the development of deep learning, metaphor detection approaches were increasingly established by neural network models, treated as a sequential tagging task. For the task of metaphor identification \cite{leong2018report}, \citet{bizzoni2018bigrams} utilized BiLSTM as main model, while \citet{wu2018neural} combined BiLSTM with CNN which achieved the best performance.

With advent of deep learning approaches, \citet{gao2018neural} used BiLSTM models based on GloVe \cite{pennington2014glove} and ELMo word vectors \cite{peters2018deep} to detect metaphoric verbs. Inspired by the linguistic theories, MIP \cite{semino2007mip, steen2010method} and SPV \cite{wilks1975preferential, wilks1978making}, \citet{mao2019end} proposed two detection models consisting of BiLSTM with attention mechanisms that relied on GloVe and ELMo embeddings. Recent work on metaphor detection have also used pretrained language models \cite{su2020deepmet, gong2020illinimet}. While we focus on metaphor generation , we use \cite{devlin2018bert} to detect metaphoric verbs to create parallel data and \cite{liu2019roberta} to rescore our generated hypothesis during decoding.

\subsection{Metaphor Generation}
Some early works made contributions to use template and heuristic-based methods \cite{abe2006computational, terai2010computational} to generate “A is like B” sentences, more popularly referred to as \textit{similes}. \citet{chakrabarty2020generating} concentrated on simile generation, applying seq2seq model to paraphrase a literal sentence into a simile. Other attempts learned from the mappings of different domains and generated conceptual metaphors of pattern “A is B” \cite{hervas2007enrichment, mason2004cormet,gero2019metaphoria}. These works paid attention to the relationship between nouns and concepts to create elementary figurative expressions. 
% \citet{gero2019metaphoria} presented an interactive system for collaboratively writing metaphors with a computer. They use an open source knowledge graph and a modified Word Mover’s Distance algorithm to find a large, ranked list of suggested metaphorical connections.
\par
Recent metaphor generation works focus mainly on verbs. \citet{yu-wan-2019-avoid} proposed an unsupervised metaphor extraction method, and developed a neural generation model to generate metaphorical sentences from literal-metaphorical verb pairs. They however do not focus on literal to metaphorical sentence transfer , but generate a sentence given a metaphorical fit word. The closest to our work is that of \citet{metagen2}, who focus on building a seq2seq model, using a special mask token to mask the metaphorical verbs as input, and the original metaphorical sentences as output. However, this model face challenges in transferring the literal sentences to metaphorical ones, while maintaining the same meaning. We, on the contrary, focus on maintaining the same meaning through parallel data creation focusing on symbolism. Additionally, we incorporate a metaphor detection model as a discriminator to improve decoding during generation.

\section{Conclusion}
We show how to transform literal sentences to metaphorical ones. We propose a novel way of creating parallel corpora and an approach for generating metaphors that benefits from transfer learning and discriminative decoding. Human and automatic evaluations show that our best model is successful at generating metaphors. We further show that leveraging symbolic meanings helps us learn better abstract representations and better preservation of the denotative meaning of the input. Future directions include learning diverse conceptual metaphoric mapping using our parallel data and constraining our metaphoric generations based on particular mapping.
%SM-cr should we mention we focused on verbs? should we mention task-based evaluation? 

\section{Ethics}
Our data is collected from Reddit and we understand and respect user privacy. Our models are fine-tuned on sentence level data obtained from user posts. These do not contain any explicit detail which leaks information about a users name, health, negative financial status, racial or ethnic origin, religious or philosophical affiliation or beliefs, sexual orientation, trade union membership, alleged or actual commission of crime. 

Second, although we use language models trained on data collected from the Web, which have been shown to have issues with bias and abusive language \cite{sheng2019woman, wallace2019universal}, the inductive bias of our models should limit inadvertent negative impacts. Unlike model variants such as GPT, BART is a conditional language model, which provides more control of the generated output. Furthermore, we specifically encode writing style from a poetic corpus in our models and train on parallel data in the direction of literal to metaphorical style.
Open-sourcing this technology will help to generate metaphoric text assisting creative writing practitioners or non native language speakers to improve their writing.
We do not envision any dual-use that can cause harm for the use of our the metaphor generation system.

\bibliography{anthology,custom}
\bibliographystyle{acl_natbib}

\appendix

\section{Appendix}
\label{sec:appendix}

For retrieving commonsense symbolism of the sentences, we use the pre-trained COMET model \footnote{\url{https://github.com/atcbosselut/comet-commonsense}} and retrieve top 5 candidates for each input.
\begin{enumerate}
    \item{\textbf{No of Parameters:}} For metaphor detection at token level we use BERT-base-cased model (110M). For generation we use the BART large checkpoint (400M parameters) and use the implementation by FAIRSEQ \cite{ott2019fairseq}
     \footnote{\url{https://github.com/pytorch/fairseq/tree/master/examples/bart}}. For discriminative decoding we use RoBERTa large model (355M)
    \item{\textbf{No of Epochs:}} For metaphor detection at token level for parallel data creation we fine-tune it for 3 epochs. We fine-tune pre-trained BART for 70 epochs for MERMAID model and save best model based on validation perplexity. For discriminator we fine-tune RoBERTa-large model for 10 epoch and save the checkpoint for best validation accuracy
    \item{\textbf{Training Time:}} For metaphor detection training time is 40 minutes.Our training time is 280 minutes for BART. For discriminator we train it for 60 minutes
    \item{\textbf{Hardware Configuration:}} We use 4 RTX 2080 GPU
    \item{\textbf{Training Hyper parameters:}} We use the same parameters mentioned in the github repo where BART was fine-tuned for CNN-DM summarization task with the exception of MAX-TOKENS (size of each mini-batch, in terms of the number of tokens.) being 1024 for us. For discrminator finetuning of roberta we use same parameters as RTE task \footnote{\url{https://github.com/pytorch/fairseq/blob/master/examples/roberta/README.glue.md}}
    \item{\textbf{Decoding Strategy \& Hyper Parameters:}}For decoding we  generate metaphors from our models using a top-k random sampling scheme \cite{fan2018hierarchical}. At each timestep, the model generates the probability of each word in the vocabulary being the likely next word. We randomly sample from the k = 5 most likely candidates from this distribution.

\end{enumerate}
\end{document}